%% file: neurips_2019.tex
\documentclass{article}

\PassOptionsToPackage{numbers, compress}{natbib}
     \usepackage[final]{neurips_2019}

\usepackage[utf8]{inputenc} % allow utf-8 input
\usepackage[T1]{fontenc}    % use 8-bit T1 fonts
\usepackage{hyperref}       % hyperlinks
\usepackage{url}            % simple URL typesetting
\usepackage{booktabs}       % professional-quality tables
\usepackage{amsfonts}       % blackboard math symbols
\usepackage{nicefrac}       % compact symbols for 1/2, etc.
\usepackage{microtype}      % microtypography
\usepackage[pdftex]{graphicx}
\usepackage{subcaption}
\usepackage[justification=centering]{caption}
\usepackage[english, nameinlink, noabbrev]{cleveref}
\usepackage[inline]{enumitem}
\usepackage{graphicx}
\usepackage{subcaption}
\usepackage{multirow}
\usepackage{datetime}
\usepackage{comment}
\usepackage{xcolor}

\usepackage{pifont}% http://ctan.org/pkg/pifont
\newcommand{\cmark}{\ding{51}}%
\title{Recurrent Instance Segmentation\\using Sequences of Referring Expressions}
\author{
  Alba Maria Hererra-Palacio\\
  Universitat Politecnica de Catalunya\\
  \And
  Carles Ventura \\
  Universitat Oberta de Catalunya\\
  \texttt{cventuraroy@uoc.edu} \\
  \And
  Carina Silberer\\
  Universitat Pompeu Fabra\\
  \texttt{carina.silberer@upf.edu} \\
  \And
  Ionut-Teodor Sorodoc \\
  Universitat Pompeu Fabra\\
  \texttt{ionutteodor.sorodoc@upf.edu} \\
  \And
  Gemma Boleda\\
  Universitat Pompeu Fabra\\
  \texttt{gemma.boleda@upf.edu} \\
  \And
  Xavier Giro-i-Nieto\\
  Universitat Politecnica de Catalunya\\
  \texttt{xavier.giro@upc.edu} \\
}

\begin{document}

\maketitle

\begin{abstract}
The goal of this work is to segment the objects in an image that are referred to by a sequence of linguistic descriptions (referring expressions). We propose a deep neural network with recurrent layers that output a sequence of binary masks, one for each referring expression provided by the user. The recurrent layers in the architecture allow the model to condition each predicted mask on the previous ones, from a spatial perspective within the same image. Our multimodal approach uses off-the-shelf architectures to encode both the image and the referring expressions. The visual branch provides a tensor of pixel embeddings that are concatenated with the phrase embeddings produced by a language encoder. Our experiments on the RefCOCO dataset for still images indicate how the proposed architecture successfully exploits the sequences of referring expressions to solve a pixel-wise task of instance segmentation.
%reduce the stress of individuals with dementia and the burden on their caregivers

\end{abstract}

\input{1_introduction}
\input{3_method.tex}

\input{4_experiments.tex}
\input{5_conclusions.tex}
\input{6_acks.tex}

\bibliographystyle{plain}
\bibliography{ref}

\end{document}

%% file: 1_introduction.tex
\section{Introduction}

\begin{figure}[!t]
    \centering
    \begin{subfigure}[t]{0.23\linewidth}
		\includegraphics[width=\linewidth]{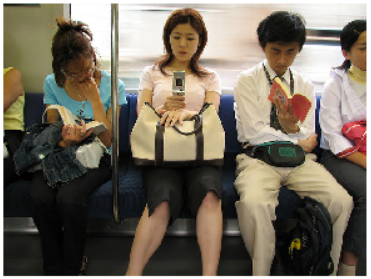}
		\caption{Input image}
	\end{subfigure}
	~
	\begin{subfigure}[t]{0.23\linewidth}
		\includegraphics[width=\linewidth]{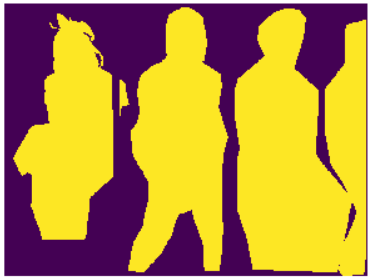}
		\caption{Of class \textit{"person"}}
	\end{subfigure}
	~
	\begin{subfigure}[t]{0.23\linewidth}
		\includegraphics[width=\linewidth]{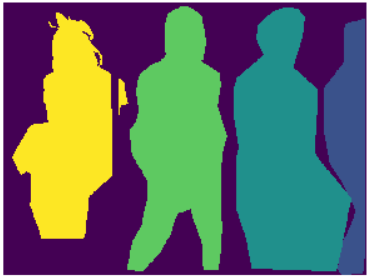}
		\caption{Of class \textit{"person"}}
	\end{subfigure}
	~
	\begin{subfigure}[t]{0.23\linewidth}
		\includegraphics[width=\linewidth]{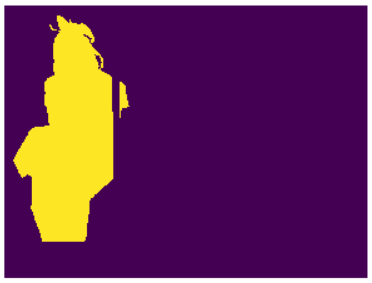}
	    \caption{Referring expression \textit{"left woman in blue"}}
	\end{subfigure}
    \caption{Comparison between different segmentation tasks: (b) object segmentation, (c) object instance segmentation and (d) segmentation from natural language expressions.}
    \label{fig:instance_seg}
\end{figure}

In this work, we tackle object instance segmentation with natural language expressions, a challenging problem with implications in the fields of computer vision and natural language processing.
The goal is to segment the referent, i.e., the target object referred to by a referring expression, in an image.
For instance, given the image in Figure \ref{fig:instance_seg}(a) and the referring expression ``left woman in blue'', the model needs to output the mask for the relevant person (Figure \ref{fig:instance_seg}(d)).
%We present an end-to-end trainable neural network that recurrently segment target objects by linguistic referring expressions. In other words, given an image and a referring expression for each of the instances to be segmented, it generates the sequence of pixel-level masks of the referents. 
Instance segmentation with referring expressions can be understood as an extension of semantic instance segmentation, where a binary mask and a categorical label are assigned to each object in an image (see comparison in Figure \ref{fig:instance_seg}).
% GBT: I don't think the following is necessary. 
%The referring expressions used for this task can take any form of linguistic description. They can contain appearance attributes (e.g. \textit{"blue"}), actions (e.g.\textit{ "sitting"}), relative relationships (e.g. \textit{"left"}), etc., and are not limited to object categories, as it is important to be able to distinguish between instances of objects without ambiguities (see comparison in Figure \ref{fig:instance_seg}). 
Humans use referring expressions to talk about objects in the world; therefore, the ability to ground referring expressions in images can be very useful in human-computer interaction scenarios, too. 

Work in this area \cite{hu2016segmentation, li2018referring, shi2018key,ye2019cross} separately represents the linguistic expression and the input image, typically using recurrent neural networks (RNN) and convolutional neural networks (CNN), respectively. Afterwards, in order to obtain a pixel-wise segmentation mask, both representations are combined and further processed.
In the case of multiple referring expressions over the same image, each of them is processed separately.
More details about related work are contained in the supplementary material.
We focus on the novel scenario in which a user does not provide a single referring expression, but a sequence of them, one for each referent.
For each expression in the sequence, our model predicts a visual grounding conditioned by not only the current reference, but also the previous ones.
%\xgn{This sentence has been replaced by the previous after the discussion by e-mail}.Our model ground each expression in a sequence is capable of conditioning the visual grounding of the current expression to be coherent with previous groundings of other referents; 
%\xgn{As we actually explore prove this hypothesis, I have removed it from the text}our hypothesis is that introducing the dependence between the referring expressions will help the model produce better quality masks, for instance preventing the prediction of a mask for a given object twice.
In addition, our model is end-to-end trainable and does not require any visual post-processing as in MAttNet  \cite{yu2018mattnet}, which was based on the Mask R-CNN  computer vision model for instance segmentation \cite{he2017mask}. Mask R-CNN, and other similar solutions, predicts a large amount of instances which are later filtered. % or ranking of object candidates

%\gbt{add ``as was done/necessary in previous work'' plus the relevant reference(s). Also, it's not clear to me what non-maximum suppression or ranking of object candidates are; either explain or move one level up in abstraction so it's understandable.}

The proposed architecture consists of:
\begin{enumerate*}[label=(\roman*)]
\item  a vision encoder, which extracts visual features of a frame,
\item a language encoder, which adds linguistic information to the model by using a pre-trained natural language processing model to extract language features for the referring expressions (phrases), and
\item a recurrent segment decoder, which uses the image and phrase embeddings from the vision and language encoders, respectively, to generate the pixel-level masks of the target objects.
\end{enumerate*} 

%The visual encoder and decoder are inspired by RSIS, an existing Recurrent network for object instance segmentation~\cite{salvador2017recurrent}. The embeddings for the referring expressions use the language encoder BERT~\cite{devlin2018bert}. Our model is trained and evaluated the RefCOCO~\cite{yu2016modeling} dataset, which provides images with pixel-level segmentation masks along with multiple referring expressions for each referent.

%% file: 3_method.tex
\section{Method}
\label{sec:method}

\begin{figure}[!t]
    \centering
    \includegraphics[width=1\linewidth]{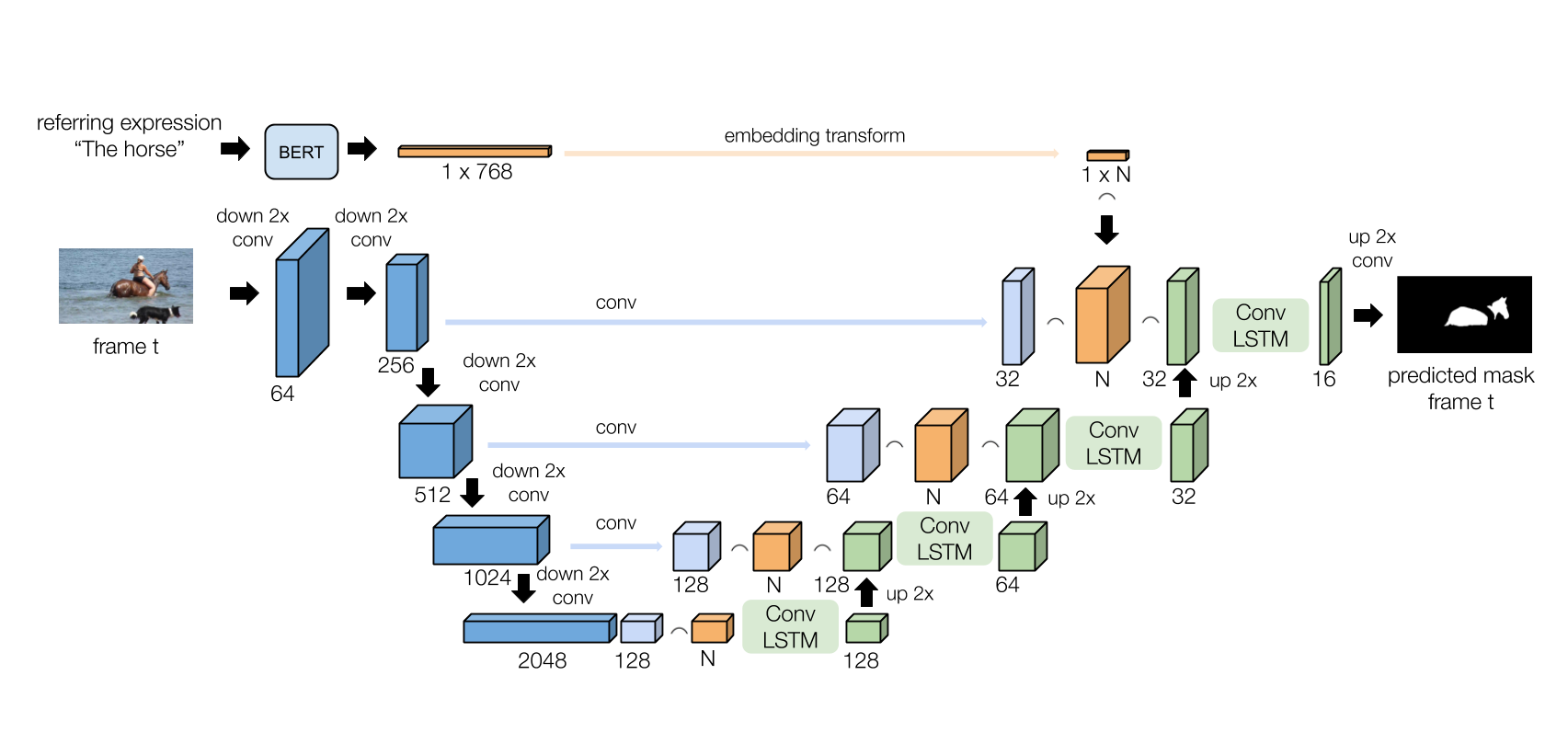}
    \caption{Our proposed recurrent architecture for recurrent instance segmentation with linguistic referring expressions. The figure illustrates a single forward pass, predicting only the mask of one instance for an image.}
    \label{fig:5}
\end{figure}

We propose an end-to-end trainable deep neural network to recurrently segment target objects indicated by linguistic referring expressions (RE).
The proposed architecture is depicted in \Cref{fig:5}.
The visual encoder and decoder are inspired by RSIS~\cite{salvador2017recurrent}, a deep neural network for object instance segmentation. The language embeddings for the referring expressions are obtained from the BERT encoder~\cite{devlin2018bert}. 
The pixel and phrase embeddings are concatenated and fed to a binary mask visual decoder.
Given a sequence of linguistic referring expressions, the recurrent nature of the mask decoder allows to condition the current prediction on the previous ones.

%, as illustrated in \Cref{fig:5}.
%The output is a pixel-level mask for each referent.
%Our model is trained and evaluated the RefCOCO~\cite{yu2016modeling} dataset, which provides images with pixel-level segmentation masks along with multiple referring expressions for each referent.

%Two encoders (vision and language) whose 
%\Cref{fig:refcoco} shows an example of the linguistic and visual inputs for a referent, and an expected output.

%The architecture, depicted in \Cref{fig:5}, is an extension of RSIS~\cite{salvador2017recurrent}, a deep neural network trained end-to-end to generates a sequence of segments with the object instances of a still image. The network is given a RE for each of the target instances (referents)  and an RGB-image, which are fed to two encoders (vision and language) to obtain pixel and phrase embeddings. Their concatenation is fed to a binary mask visual decoder which outputs a pixel-level mask for each referent.% (see \Cref{fig:refcoco} for an example).

The BERT~\cite{devlin2018bert} encoding, represented at the top branch of \Cref{fig:5}, is used without any fine-tuning.
%is used to obtain representations for the referring expressions without fine-tuning any of the parameters. , i.e, inputs to our language encoder, as seen at . 
%By extracting the activations from one or more layers we obtain a sequence of contextualized token embeddings (hidden states) and a pooled output (phrase embedding), as illustrated in \Cref{fig:bert2}. 
We use the base model of 12 encoder layers (transformer blocks), 768 hidden units and 12 attention heads.\footnote{Publicly available as \texttt{bert-base-cased model} at  \url{https://github.com/huggingface/pytorch-transformers}.}
%At the output of the last layer of the model, a set of contextualized embeddings of the words (hidden states for this model) is stored in a four-dimensional tensor, in the following order \gbt{is the order relevant? }: layer number (12 layers/hidden states), batch number (1 sentence) \gbt{I don't understand ``1 sentence'' here}, word (token) number in each sentence, and hidden unit/feature number (768 features). 
Given a referring expression, BERT outputs a set of contextualized embeddings which comprises the hidden states of each encoder layer of each word (token).  
We average the last hidden layer of each token producing a single 768 length vector for each referring expression. 
To avoid memory problems while training the model and to balance the dimensions of the language and visual embeddings, we reduce the dimensionality of the textual embeddings to 64 with principal component analysis (PCA)~\cite{pca}.

The visual encoding and decoding schemes were adopted from the RSIS~\cite{salvador2017recurrent} model for semantic instance segmentation.
The input image is encoded with a ResNet-101~\cite{he2016deep} model pre-trained on ImageNet~\cite{imagenet_cvpr09}.
The ResNet architecture is truncated at the last convolutional layer, thus removing the last two layers (pooling layer and classification layer). In contrast to the language branch, the image encoder was finetuned for the task. The output of each convolutional block is used as an image feature, which provides a set of visual features at different resolutions, as shown in dark blue at the left of \Cref{fig:5}. 
%We want to obtain image representations from the input frame that allow a precise pixel segmentation of the instances in it. So, for a given frame, We use the encoder architecture used by 
 
The input to the mask decoder for a given referent consists of a set of multi-resolution pixel embeddings obtained by the visual encoder, and the phrase embedding provided by the language encoder. Consequently, visual features are shared among all the referents for the same image, and the output of the decoder is a sequence of object segmentation predictions, one for each referent. 

The mask decoder is an extension of the multi-resolution one proposed in RSIS~\cite{salvador2017recurrent}.
In order to keep the inherent spatial information in the visual features when segmenting an instance, for each resolution, we concatenate the corresponding language embedding to each feature map along the channels' dimensions (depth) of the visual tensors. This allows every pixel embedding to receive the whole representation of the language information.
%combine the language embeddings with multi-resolution visual features in the decoder, also inher.
The ConvLSTM~\cite{xingjian2015convolutional} layers used in the decoder allows to condition the predicted masks with those masks predicted for previously presented referring expressions over the same image. %The ConvLSTM~\cite{xingjian2015convolutional} layers used in the decoder contain a state memory that has the potential to remember previous predictions.

%which uses ConvLSTM~\cite{xingjian2015convolutional} layers for the different resolutions of the input features (concatenated visual and language embeddings) as shown in \Cref{fig:5}, is adequate to retrieve segmentation pixel-masks for a referent.
%Although this project is focused on the segmentation in images, videos can be segmented using this decoder architecture.

Similarly to~\cite{salvador2017recurrent}, the cost function is defined as the soft Intersection over Union score between the predicted mask and the ground truth mask for a given referent.
Since we do sequential processing, during training we use as ground truth the mask corresponding to the referring expression being processed at each timestep.
%\gbt{In the following sentence, I don't understand "assignment" in "assignment between predicted masks and ground truth". Do you mean "comparison", "matching", "alignment", "evaluation of predicted masks compared to the ground truth?".
%Alternative formulation:
%"Since we do sequential processing, during training we use as ground truth the mask corresponding to the referring expression being processed at the relevant timestep."}
%During training, the assignment between predicted masks and ground truth is based on the order in which the referring expressions for each referent are processed by the model. 

%% file: 4_experiments.tex
\section{Experiments}
\label{sec:experiments}

%This section presents the results obtained with the proposed architecture for instance segmentation with linguistic referring expressions on the RefCOCO dataset~\cite{yu2016modeling}. %So the experiments, which are described below, are limited to the spatial model.
The experiments show how the introduction of the referring expression encoder successfully conditions the mask to predict, and that the order of the phrases within the input sequences affects the performance of the model. The experiments have been performed on the RefCOCO dataset~\cite{yu2016modeling}, a dataset with 142K referring expressions for 50K objects in 20K images from MSCOCO \cite{lin2014microsoft}, that is, where the target objects are of 80 common categories.
More details on the dataset and model training are contained in the supplementary material.

We validate the performance of the referring expression branch by comparing the results with the baseline case of not using referring expressions.
In this case, RSIS is used to generate a fixed-length sequence of instance masks. The length of the sequence is always larger than the amount of reference phrases associated with the image, avoiding to penalize the segmentation of objects for which no referring expression is presented to the model.
Instead of forcing a specific order when matching the predicted masks and ground truth masks, the Hungarian algorithm~\cite{kuhn1955hungarian} carries out an optimal assignment between them using the soft Intersection over Union score as cost function.
The results presented in \Cref{table:results_images} show four different configurations in terms of referent order and batch size, with (our multimodal model) or without (RSIS, i.e. the visual model) referring expressions.
Our solution consistently outperforms RSIS, even when RSIS is completely free to generate its masks in any order.
%the configuration without referring expression.
These results show that the linguistic phrases are successfully used to identify the right target instance, and that, in addition, the quality of the masks actually improves over the language-free task addressed by RSIS.
\begin{table}[!b]
	\centering 
	\caption{Results on RefCOCO with and without referring expressions.}
	\label{table:results_images}
	\begin{tabular}{lcccccccc}
		\toprule
		\textbf{Referent order}         & \textbf{Batch size} & \textbf{Referring} &        \multicolumn{3}{c}{Instance IoU $\uparrow$}        &        \multicolumn{3}{c}{Overall IoU $\uparrow$}         \\
		                                &                     &                       \textbf{expression}        & \textbf{val} & \textbf{testA} & \textbf{testB} & \textbf{val} & \textbf{testA} & \textbf{testB} \\ \midrule
		\multirow{4}{*}{By area} &         128         &                               & 21.82        & 25.56          & 18.86          & 18.48        & 21.27          & 16.48          \\
		                                &         128         & \cmark                        & 26.08        & 29.63          & 22.81          & 23.67        & 26.47          & 21.13          \\
		                                &         32          &                               & 21.68        & 23.50          & 19.67          & 19.42        & 21.02          & 17.94          \\
		                                &         32          & \cmark                        & 26.12        & 28.66          & 23.82          & 23.88        & 25.81          & 22.23          \\ \midrule
		\multirow{5}{*}{Random}         &         128         &                               & 20.36        & 22.70          & 15.78          & 17.65        & 19.32          & 15.22          \\
		                                &         128         & \cmark                        & 27.54        & 31.45          & 24.39          & 24.75        & 27.76          & 22.26          \\
		                                &         32          &                               & 20.13        & 23.13          & 19.04          & 17.77        & 19.83          & 17.24          \\
		                                &         32          & \cmark                        & 39.79        & 45.31          & 34.04          & 35.70        & 40.28          & 31.28          \\ 
		                                & 16                  & \cmark                        & \textbf{42.66} & \textbf{47.48 } & \textbf{37.51} & \textbf{36.95} & \textbf{41.42} & \textbf{32.72} \\

		                                \bottomrule
	\end{tabular}
\end{table}

 We also investigated the effect of the order within the sequence of referring expressions used to train the model. We considered two options:
\begin{enumerate*}[label=(\roman*)]
        \item by area,
        \item randomly.
\end{enumerate*}
 The results in \Cref{table:results_images} indicate that the best strategy is to randomly feed the referents and use small batch sizes.
 %\gbt{I rephrased the following sense, please check that I got it right:}
 The fact that our best result is obtained for the smallest batch size (16) and a random ordering may indicate that our model overfits and that further reducing the amount of parameters to learn may even increase the performance.
 %\gbt{why not dropout?}
 If we focus on the batch size\ 32 with the referring expression, we can also observe that the random configuration almost doubles the performance with respect to training with objects sorted by area. 
 These results highlight the importance to randomize the training  expressions to avoid learning undesirable data biases.

\begin{figure}[!t]
    \centering
    \begin{subfigure}[t]{0.18\linewidth}
	    \includegraphics[width=2.7cm, height=3.6cm]{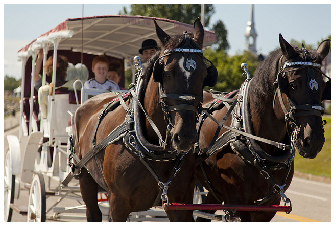}
	    \caption{}
	\end{subfigure}
	~
	\begin{subfigure}[t]{0.18\linewidth}
		\includegraphics[width=2.7cm, height=3.6cm]{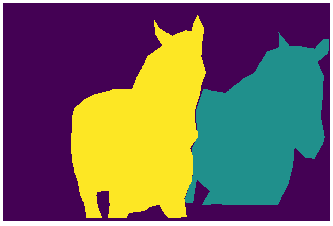}
		\caption{}
	\end{subfigure}
	~
	\begin{subfigure}[t]{0.18\linewidth}
		\includegraphics[width=2.7cm, height=3.6cm]{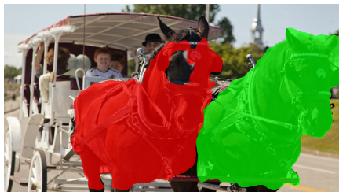}
		\caption{}
	\end{subfigure}
	~
	\begin{subfigure}[t]{0.18\linewidth}
		\includegraphics[width=2.7cm, height=3.6cm]{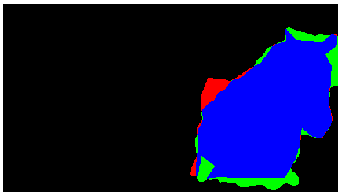}
		\caption{\textit{"right horse"}}
	\end{subfigure}
	~
    \begin{subfigure}[t]{0.18\linewidth}
		\includegraphics[width=2.7cm, height=3.6cm]{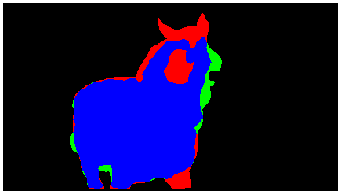}
		\caption{\textit{"left horse"}}
	\end{subfigure}
	
    \caption{Qualitative results: (a) original image, (b) ground truth, (c) segmentation result, (d) and (e) pixel-wise predictions (red pixels are false negatives, blue true positives and green false positives).}
    \label{fig:resultsB}
\end{figure}

\Cref{fig:resultsB} shows some qualitative results generated by our network. 
The depicted results are among the good predictions of the algorithm and show how our model can distinguish between different instances of the same class.

\begin{figure}[!t]
    \centering
    \begin{subfigure}[t]{0.23\linewidth}
	    \includegraphics[width=3cm, height=4cm]{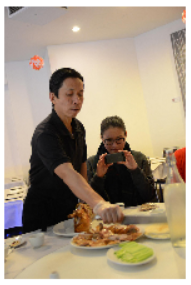}
	    \caption{}
	\end{subfigure}
	~
	\begin{subfigure}[t]{0.23\linewidth}
		\includegraphics[width=3cm, height=4cm]{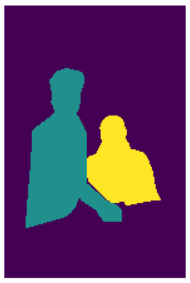}
		\caption{}
	\end{subfigure}
	~
	\begin{subfigure}[t]{0.23\linewidth}
		\includegraphics[width=3cm, height=4cm]{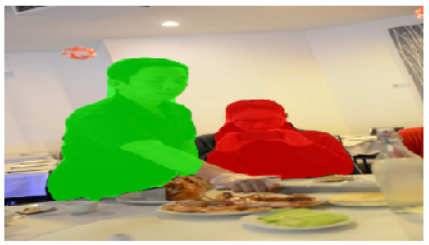}
		\caption{\textit{"man on the left"}, \textit{"right gal"}}
	\end{subfigure}
	~
	\begin{subfigure}[t]{0.23\linewidth}
		\includegraphics[width=3cm, height=4cm]{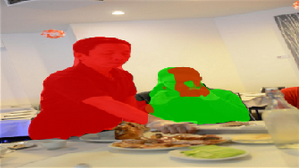}
		\caption{\textit{"right gal"}, \textit{"man on the left"}}
	\end{subfigure}

    \caption{Changing referring expressions order: (a) original image, (b) ground truth, (c) segmentation result (original order), and (d) segmentation results (inverse order).}
    \label{fig:results_comp}
\end{figure}

Finally, \Cref{fig:results_comp} depicts how the order of the segmented objects is consistent with the order of the referring expressions. By reversing the order of the phrases, the order of the generated masks is also reversed, which shows that the model has learnt how to associate REs with visual objects. 
Note that the generated masks are not exactly the same. 
This indicates that the model indeed conditions its segmentation decisions on  its predictions for previous REs.
Note how the generated masks are not exactly the same, another evidence that suggests that the order affects the segmentation results.
%\gbt{Not really; it is evidence that the model has learnt how to associate REs with objects, but not that the sequential order improves segmentation quality.}
%\xgn{I replaced 'segmentation quality' for 'segmentatino results' as proposed by Gemma}

%% file: 5_conclusions.tex
\section{Conclusions}
\label{ref:conclusions}
    
This work has proposed a solution for visual object segmentation by adding a new linguistic branch to the RSIS deep neural architecture.
The concatenation of the phrase embeddings of the referring expression to each pixel embedding of the RSIS decoder has the potential to successfully condition the predicted mask to the desired object.
The recurrent nature of the decoder allows to process sequences of referring phrases and condition the output based on the previous predictions.
%allows selecting which instance to segment in a weakly supervised scenario.
%While our quantitative results do not exceed the state of the art in terms of accuracy,
The proposed architecture is trained end-to-end and avoids the additional computation required by post-processing steps such as non-maximum suppression or ranking of object proposals.
Further details and qualitative results are contained in the supplementary material
\footnote{\url{https://vigilworkshop.github.io/static/papers/30_supp.pdf}}.

%% file: 6_acks.tex
\section*{Acknowledgements}

\textbf{UPC:} This work has been developed in the framework of project TEC2016-75976-R, funded by the Spanish Ministerio de Economía y Competitividad and the European Regional Development Fund (ERDF), and the Industrial Doctorate 2017-DI-011 funded by the Government of Catalonia. We gratefully acknowledge the support of NVIDIA Corporation with the donation of some of the GPUs used for this work.

\textbf{UOC:}This work has been partially supported by the Ministerio de Economia, Industria y Competitividad (Spain), under the Grant Ref. RTI2018-095232-B-C22.

\textbf{UPF:}This project has also received funding from the European Research Council
(ERC) under the European Union's Horizon 2020 research and innovation
programme (grant agreement No 715154),
and from the Ram\'on y Cajal programme (grant RYC-2015-18907).
We gratefully acknowledge the support of NVIDIA Corporation with the donation of GPUs used for this research, and the computer resources at CTE-POWER and the technical support provided by Barcelona Supercomputing Center (RES-FI-2018-3-0034).
This paper reflects the authors' view only, and the EU is not
responsible for any use that may be made of the information it
contains.